\documentclass{bmvc2k}

\title{AccioScene: Compositional 3D Scene Generation via Graph Diffusion and Interaction-driven Critics}

\addauthor{
\parbox{0.95\textwidth}{
\centering
Yao Wei$^{1,2*}$ \quad
Matteo Toso$^{2}$ \quad
Pietro Morerio$^{2}$ \quad
Changjae Oh$^{1}$\\
Michael Ying Yang$^{3}$ \quad
Alessio Del Bue$^{2}$\\[0.6em]
{\small
$^{1}$ Queen Mary University of London, UK\\
$^{2}$ Italian Institute of Technology (IIT), Italy\\
$^{3}$ University of Bath, UK
}
}
}{}{1}

\addinstitution{}

\runninghead{Wei et al.}{AccioScene}

\def\eg{\emph{e.g}\bmvaOneDot}

\begin{document}

\maketitle

\begin{abstract}
This paper presents a framework for generating 3D indoor scenes from text prompts. Existing methods often formulate scene synthesis as an object layout prediction problem conditioned on a single input modality, such as a text description, room shape, or scene graph. This design can lead to object collisions and limited functional plausibility, reducing its practical applicability. To address these limitations, we introduce a multi-stage pipeline that better reflects practical scene creation scenarios. Given a text prompt describing partial scene content, our method first uses graph diffusion to produce a contextually coherent scene graph and then predicts a realistic object layout. In addition, we incorporate lightweight human-object interaction priors to encourage human-centric and functional arrangements, with explicit spatial constraints to reduce interpenetration. Our approach generates coherent 3D scenes with viable layouts that better support human interaction. Experiments on the 3D-FRONT dataset demonstrate that our method achieves competitive or state-of-the-art performance compared with existing approaches, while improving the physical plausibility of generated scenes.
\end{abstract}

\section{Introduction}
\label{sec:intro}

Generative models for scene synthesis are rapidly transforming the way 3D environments are produced, edited, and explored. By generating object layouts and appearances from user instructions, they eliminate the complexity and time cost of manual scene creation. This capability has immediate applications in AR/VR~\cite{VRCopilot2024,gao2022get3d,lin2023magic3d} and strong potential to revolutionize creative workflows for artists and designers~\cite{tseng2020modeling,Canvas2024}, while also serving as a foundation for training models that reason and act in 3D spaces. Despite recent progress, 3D scene generation still faces substantial challenges. Methods must not only produce spatial layouts that are visually appealing, coherent, and free of artifacts, but also ensure physical and functional realism. For practical applications, scenes must respect physical constraints, support intended human interactions, and consider human accessibility and affordances.

\begin{figure}[tb]
  \centering
  \includegraphics[width=.9\columnwidth]{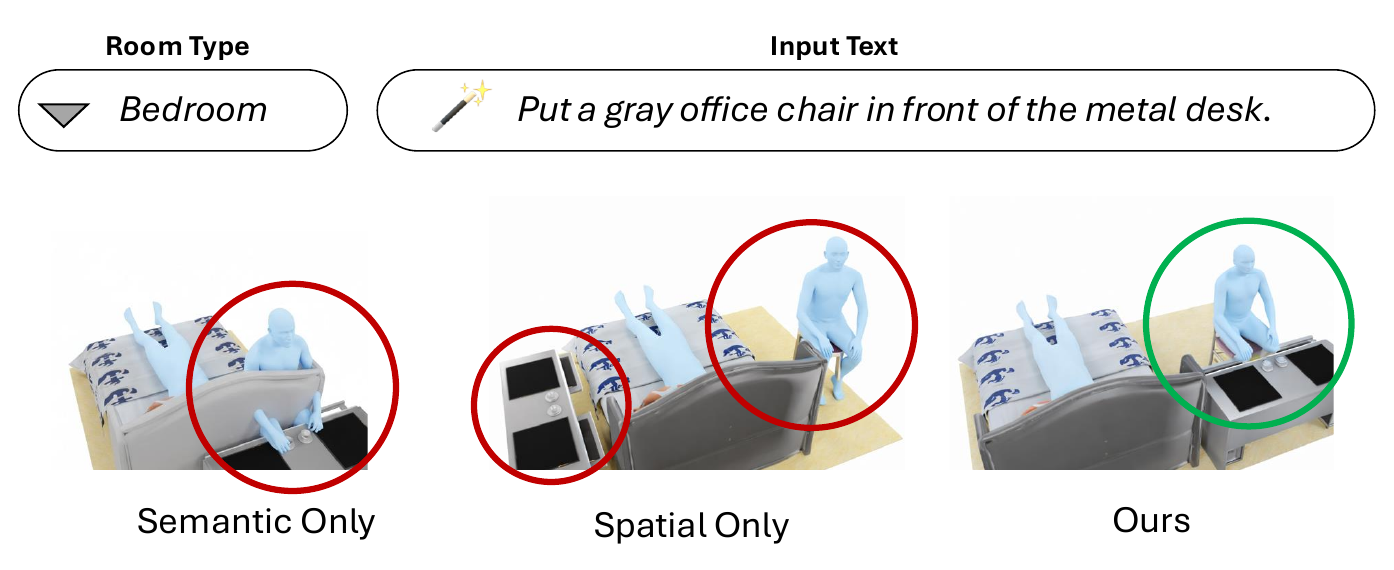}
  \caption{Naively estimating layouts from semantic only descriptions causes visual artifacts such as object collisions, while simply adding spatial constraints results in functionally incoherent arrangements. In this work, we propose \textbf{\modelname{}} which integrates gr\textbf{A}ph diffusion techniques and interact\textbf{IO}n-driven \textbf{C}ritics to populate the scene with semantically coherent objects, while simulating human-object interactions to produce spatially plausible layouts.
  }
  \label{fig:example}
\end{figure}

Existing approaches can be grouped according to their input modality in Semantic-driven or Spatial-driven methods. The former include Text-to-Scene (T2S) methods, which use detailed textual descriptions~\cite{tang2024diffuscene,lin2024instructscene}; and Graph-to-Scene (G2S) methods, which rely on complete scene graphs~\cite{zhai2024echoscene}. The latter include Floor-to-Scene (F2S), which take 2D floor plans~\cite{paschalidou2021atiss}; and Human-to-Scene (H2S), which use human motion sequences. Previous G2S and H2S methods~\cite{zhai2024echoscene,wei2025planner3d,yi2023mime,tang2024diffuscene} are often impractical, requiring fine-grained scene graphs or well-aligned motion-scene pairs, making them time-consuming and memory-intensive. Natural language instructions are more intuitive, but T2S methods~\cite{zhou2024gala3d,ocal2024sceneteller} typically depend on explicit long descriptions. Recent work such as \textit{InstructScene}~\cite{lin2024instructscene} handles implicit or short text prompts, but risks under-specification, leading to unrealistic details or missing key objects. F2S methods suffer from similar omissions. Most of these methods treat layout estimation as a regression task, overlooking functional aspects.

To overcome these limitations, we introduce \modelname{}, a framework that understands user intent in a compositional way. Our method enriches sparse textual descriptions with human-object interaction, to provide a contextual cue. This minimal yet realistic set of inputs is used to generate a partial scene graph encoding object positions, styles, room type, and associated actions. Two graph diffusion models expand the graph with contextually appropriate objects. The scene is then assembled by retrieving matching 3D assets and refined via human-object interaction modeling, ensuring both semantically coherence and spatial plausibility. This pipeline enables realistic indoor scene synthesis from implicit, low-burden inputs while improving generalization to real-world scenarios. As shown in Section~\ref{sec:exp}, it produces coherent and functionally complete rooms, surpassing the T2S and L2S baselines. We do not compare against Humans-to-Scene (H2S) methods, as conditioning a scene on full human trajectories and actions is an overly constrained problem that requires information difficult to obtain in most practical applications and differs fundamentally from the other input modalities considered.

Our \textbf{key contributions} can be summarized as follows:
\begin{itemize}
    \item  We propose \modelname{}, a compositional 3D indoor scene synthesis framework that integrates scene graph diffusion with interaction-aware semantic and spatial critics for generating scenes from implicit text instructions.
    \item Our method introduces lightweight human-object interaction cues into both semantic reasoning and spatial assembly, enabling self-correction of object placement and improving functional and physical plausibility.
    \item We conduct extensive experiments on 3D-FRONT, showing that \modelname{} achieves competitive or state-of-the-art performance in semantic coherence, visual fidelity, and physical plausibility under low-burden user inputs.
\end{itemize}

\section{Related Work}
\label{sec:related}

\subsection{3D Scene Generation} 

Recent advances in object-centric 3D generation have demonstrated remarkable progress in synthesizing individual 3D objects with diverse geometry and rich texture details~\cite{poole2023dreamfusion}. Approaches based on generative models~\cite{lin2025partcrafter} and large-scale 3D datasets~\cite{deitke2023objaverse} have significantly improved the fidelity and diversity of generated objects. However, extending these techniques to multi-object scene synthesis remains challenging. Unlike single-object generation, scene generation requires compositional modeling across multiple objects, where semantic consistency, spatial relations, and functional compatibility must be jointly considered.

The goal of indoor scene generation is to synthesize reasonable furniture arrangements that satisfy specific room types within a 3D space. Early works~\cite{ritchie2019fast,wang2021sceneformer,paschalidou2021atiss} typically employ CNN- or transformer-based autoregressive generative models to progressively estimate the 3D layout, including object position, size, and orientation. However, these approaches~\cite{paschalidou2021atiss,feng2024layoutgpt} often rely on coarse spatial conditions, such as floor plans, which provide limited control over scene content and object semantics. Subsequent research introduces richer structural representations such as scene graphs~\cite{zhai2024echoscene,wei2025planner3d,GraphDreamer}, where nodes represent objects and edges encode relationships. While scene graphs provide stronger compositional priors, manually designing graph structures remains non-trivial, particularly when dealing with diverse object categories and complex inter-object relations. More recently, text-based scene generation has attracted increasing attention~\cite{tang2024diffuscene,lin2024instructscene,bai2025freescene}. With the rise of large language models (LLMs), several works attempt to use LLMs for open-vocabulary scene generation~\cite{feng2024layoutgpt,ocal2024sceneteller,zhou2024gala3d,fu2024anyhome}. Although these methods enable flexible text-driven scene specification, the results often suffer from hallucination and lack precise spatial or structural control. In parallel, some pioneer work adopts agentic frameworks~\cite{ccelen2024design,yang2025sceneweaver,xia2026sage}, where multiple specialized agents collaborate to accomplish scene generation and manipulation. However, they are computationally expensive and may suffer from error accumulation across cascaded modules.

Despite these advances, controllable and practical 3D scene generation remains an open problem. Existing approaches either rely on limited structural inputs that provide insufficient semantic control, or require complex conditions such as detailed graphs or human interaction, which are costly to obtain and difficult to scale. These limitations motivate us to revisit the problem from a different perspective: designing a user-friendly conditioning mechanism that allows intuitive scene specification while enabling the system to iteratively refine and optimize the generated scene.

\subsection{Human-Object Interaction} 
Synthesizing humans interacting with the 3D environment is a key challenge for advancing 3D scene understanding applications. Recent developments in 3D human-scene interaction have been separately investigated in two distinct areas: human-to-scene synthesis and scene-to-human synthesis. On one hand, human-aware scene synthesis methods such as Pose2Room~\cite{nie2022pose2room}, SUMMON~\cite{ye2022scene}, MIME~\cite{yi2023mime}, and SHADE~\cite{hong2024human} generate 3D scenes from floor plans and human motion sequences. However, these approaches require human motion sequences during inference, which are not always easily accessible. On the other hand, most attempts on scene-aware human synthesis have been made to generate human poses and motions given a 3D scene. SceneGrok~\cite{savva2014scenegrok} predicts action maps from scanned 3D scenes. POSA \cite{hassan2021populating} introduced an ego-centric representation grounded in the surface-based 3D human model SMPL-X \cite{pavlakos2019expressive}, incorporating contact labels and scene semantics. Dynamic human-object interactions have also been recently explored, e.g., text-to-motion synthesis \cite{li2024controllable,yi2024generating}. However, these methods rely on precise 3D scenes, and a noisy scene mesh can cause penetration between the human body and scene, and increasing computational complexity. Instead, we introduce scene synthesis through optimization, which utilizes the inferred 3D human models to enhance the functionality of synthesized 3D scenes.

\begin{figure*}[tb]
  \centering
   \includegraphics[width=\columnwidth]{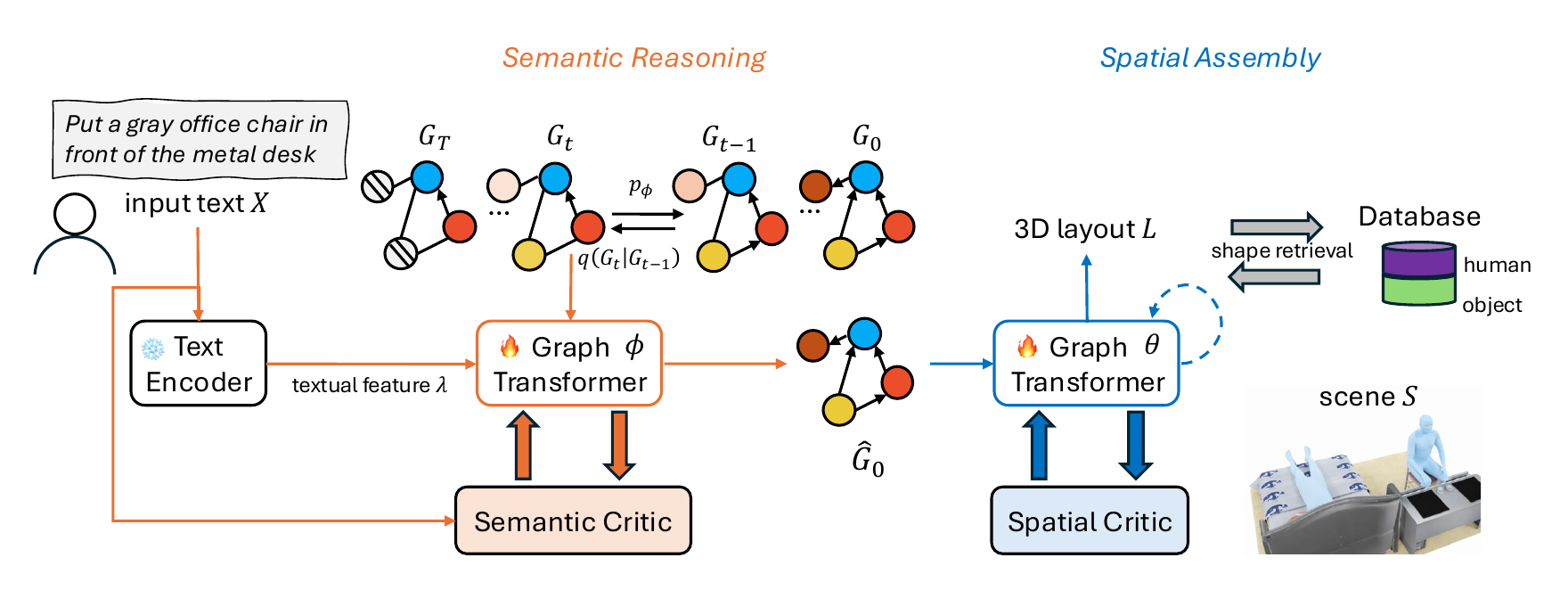}
   \caption{\textbf{Architecture of the proposed \modelname{}.} For the stage of text-to-graph generation, a graph transformer $\phi$ is trained to recover the masked graph nodes and relations, which can be used to reason about the user intent given implicit text inputs. With a reconstructed scene graph, another graph transformer network $\theta$ is trained for 3D object bounding box regression, performing graph-to-layout generation followed by shape retrieval. Two complementary interaction-driven critics provide semantic and spatial feedback, yielding 3D scenes that exhibit physical and functional plausibility.
   }
   \label{fig:pipeline}
\end{figure*}

\begin{figure}[tb]
  \centering
   \includegraphics[width=.65\linewidth]{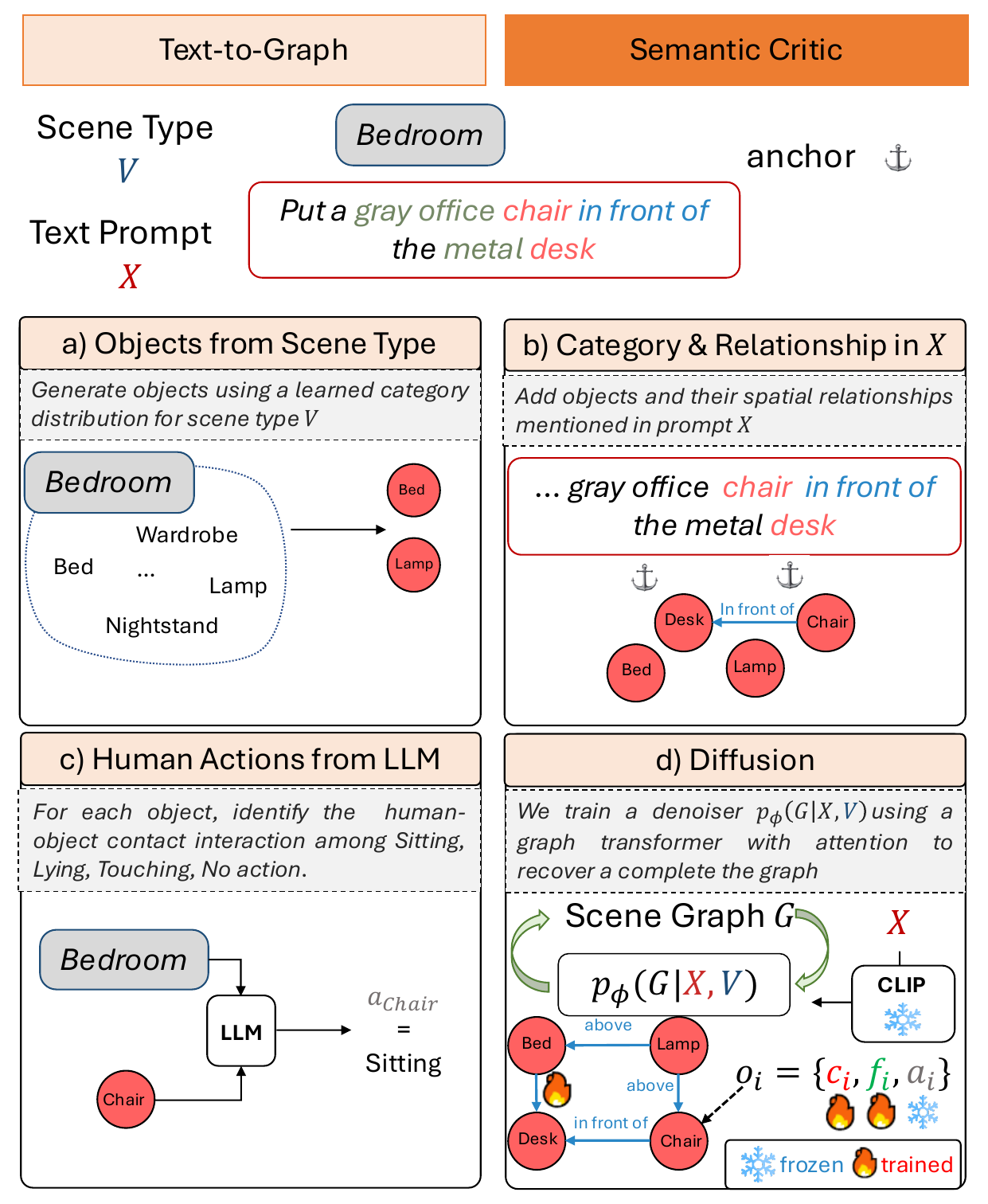}
   \caption{\textbf{Semantic reasoning.} By integrating scene-level object commonsense co-occurrence and object-level human interactions, we enrich the scene graph representations. Given the scene type $V$ and a text prompt $X$, we predict likely object categories $c$ by a) leveraging the learned distribution of objects in a collection of same-type 3D scenes and b) extracting from $X$ object categories and their spatial relationship $R$. Then, we c) use Llama to predict possible human contact interaction $a$ with the objects. d) We then use a graph diffusion network to generate the complete scene graph $G$. 
   }
   \label{fig:semantic_critic}
\end{figure}

\section{Method}
\label{sec:method}

\subsection{Problem Statement}

Instead of employing well-crafted scene graphs or detailed language descriptions as inputs, we explore easy-to-use text descriptions to guide scene generation during both training and inference, which greatly reduces manual efforts. Fig. \ref{fig:pipeline} illustrates the overall architecture. First, we semantically reason about the user intent by constructing scene graphs $G$ from text inputs $X$ with a denoising diffusion model parameterized by $\phi$. Meanwhile, a semantic critic is introduced to enrich scene graph representations by integrating scene-level object commonsense co-occurrence and object-level human interactions. Then, another denoising diffusion model parameterized by $\theta$ is trained to predict the 3D layouts $L=\{l_1, l_2, ..., l_n\}$ of the graph nodes, where $n$ is the number of object instances within the scene. In addition, a spatial critic is designed to assess whether the shapes of 3D objects and the posed human body have functional coherence in an immersive manner. At inference, given a partial textual scene description, our \modelname{} generates compositional 3D scenes with objects that are consistent with the user inputs and compliant with a set of human-object interactions. 

\subsection{Semantic Reasoning}
\label{subsec:reasoning}

Scenes are typically composed of multiple objects, where graphs $G=\{O,R\}$ can serve as effective tools for modeling scene-level object commonsense co-occurrence. The graph nodes represent objects $O=\{o_1, o_2, ..., o_N\}$, and the relations $R=\{r_1, r_2, ..., r_M\}$ are formulated by directed edges connecting nodes, where $N$ and $M$ are the number of objects and relations, respectively. Besides the lights that are attached to the ceiling, we assume objects are placed on a horizontal rectangular plane, i.e., floor plan, the size of which is automatically determined by the 2D contour of 3D object layouts. Walls and related elements are not considered.

Similar to the baseline InstructScene \cite{lin2024instructscene}, we regard implicit text inputs as incomplete or masked scene graphs $G_t$, where $t$ is a timestep in the diffusion model. To recover a complete original scene graph $G_0$, a denoiser (parameterized by $\phi$) is trained to reconstruct graphs from masked graphs. For the condition of the diffusion model, we extract textual features $\lambda$ from $X$ using a pre-trained and frozen CLIP text encoder \cite{radford2021learning}. The denoiser network is a graph Transformer consisting of graph attention, MLP layers and cross-attention modules, and it is trained to model the conditional distribution $p_\phi(G|X)=p_\phi(O,R|\lambda)=p_\phi(C,F,R|\lambda)$ by minimizing the objective,
\begin{equation}
\mathcal{L}^{G|X}_\phi=\mathcal{L}^{O,R|\lambda}_\phi=\delta_c\mathcal{L}^{C|\lambda}_\phi+\delta_f\mathcal{L}^{F|\lambda}_\phi+\delta_e\mathcal{L}^{R|\lambda}_\phi,
\label{eq:loss_diff1}
\end{equation}
where $\mathcal{L}_\phi$ represents the variational upper bound on $\mathbb{E}_{q(G_0)}[-\log p_\phi(G_0)]$, $C$ and $F$ indicate object categories and features, respectively. The object features $F$ are represented by OpenShape \cite{liu2024openshape} features quantized from a code book $Z$, derived through a pre-trained vector-quantized variational autoencoder (VQ-VAE). 

Through constructing commonsense graph from implicit text descriptions, more informative scenes can be structured by involving additional objects and relations. However, it can be observed that this one-shot text-to-graph generation scheme often leads to unsatisfactory results. For example, there is no guarantee that the objects, features and relationships specified in the text prompt are present in the reconstructed scene graphs. Moreover, artifacts may appear due to a lack of object function awareness. We therefore introduce a critic that evaluate semantic plausibility, as shown in Fig.~\ref{fig:semantic_critic}. 

On one hand, the semantic critic enforces text-graph semantic alignment. The objects and relations explicitly specified in the input text are first identified and treated as semantic anchors. These anchors act as constraints during the graph diffusion process, ensuring that the generated scene graph preserves the key conditions provided by the text prompt and preventing essential objects or relations from being omitted. On the other hand, the critic evaluates functional plausibility among objects within the reconstructed graph. While commonsense co-occurrence statistics capture frequent object configurations, they do not explicitly model functional dependencies between objects. For instance, a desk is typically associated with a chair. To address this limitation, we incorporate human–object interactions to encourage synthesized scenes that are functionally coherent. Specifically, the $i$-th graph node $o_i=\{c_i, f_i, a_i\}$ consists not only of an object category $c$, but also on possible human actions $a$, and shape and texture features $f$. The actions $a$ are inferred by Llama-2 \cite{touvron2023llama} via in-context learning. Our prompts to Llama-2 are composed of task specification, query, and in-context exemplars. The task is described using a very compact prompt: ``\textit{In a \{ROOM TYPE\}, please infer the potential human-object interactions given a list of objects}'', where \{\textit{ROOM TYPE}\} will be replaced by the queried scene type $V$. The object list serves as the query condition, and we provide a set of supporting exemplars as, for example, \textit{sofa} typically supports human \textit{sitting} or \textit{lying}. Here, since the objects are not articulated, we focus on typical human actions: \textit{sitting}, \textit{lying} and \textit{touching}, and $a$ is set to \textit{None} if human-object interactions do not exist. 

\begin{figure}[t]
  \centering
   \includegraphics[width=.8\linewidth]{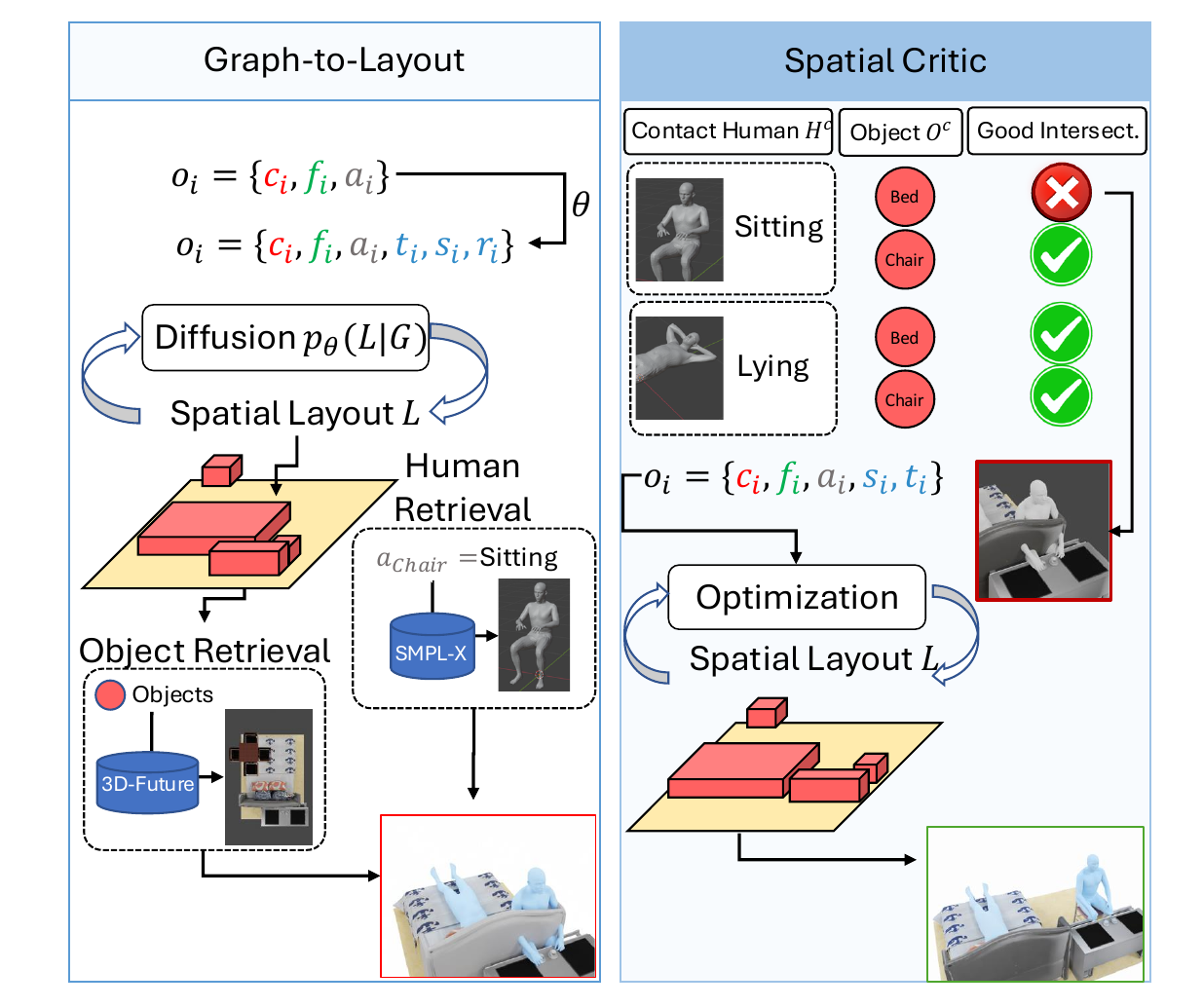}
   \caption{\textbf{Spatial assembly.} Graph-to-layout: the second diffusion model generates the spatial layout of the object, assigning to each object a translation $t$, size $s$, and orientation $r$ while leaving the previously defined features unaltered; human actions and objects' categories are used to retrieve 3D meshes and place them at poses $(r,s,t)$. Spatial critic: iterating over all predicted human-object actions, we optimize the pose of any object that would results into an intersection of the human mesh and the scene's objects. During this stage all other features are left unaltered.}
   \label{fig:spatial_critic}
\end{figure}

\subsection{Spatial Assembly}
\label{subsec:assembly}

Fig.~\ref{fig:spatial_critic} illustrates the pipeline of spatial assembly. First, we train a diffusion model parameterized by $\theta$ to predict the 3D layouts $L=\{l_1, l_2, ..., l_n\}$ of the graph nodes, where $n$ is the number of object instances within the scene. The $j$-th layout $l_j=\{t_j, s_j, r_j\}$ is parameterized by 8 values defining a translation $t\in \mathcal{R}^3$, a per-axis size $s\in \mathcal{R}^3$, and a rotation represented as $cos(r)$ and $sin(r)$. This denoiser network is also a graph Transformer, sharing the same architecture with model $\phi$ without sharing any weights. It is trained to represent the conditional distribution $p_\theta(S|G)=p_\theta(L|G)$. Model $\theta$ is optimized by minimizing the Mean Squared Error (MSE) between ground truth and predicted layouts, formulated as,
\begin{equation}
\mathcal{L}^{L|G}_\theta=\frac{1}{n}\sum_{i=1}^n(\hat{t_i}-t_i)^2+\frac{1}{n}\sum_{i=1}^n(\hat{s_i}-s_i)^2+\frac{1}{n}\sum_{i=1}^n(\hat{r_i}-r_i)^2.
\label{eq:loss_diff2}
\end{equation}

Then, textured 3D mesh models are retrieved from the 3D-FUTURE~\cite{fu20213d2} database according to the estimated object attributes including category, feature and size, i.e., $(c,f,s)$. The retrieval process starts with filtering by object categories, and sorting by feature cosine similarities. After selecting top-K textured 3D models, these selected models are sorted again by size. This way, the model with the most similar geometries and appearances can be retrieved.

Meanwhile, 3D humans are retrieved using contact object category $c$ and human action $a$, from a collection of SMPL-X body models~\cite{pavlakos2019expressive} with poses from RenderPeople scans \cite{patel2021agora}. There are five contact human bodies: sitting with hands at sides, sitting with arms on table, lying down with hands crossed behind head, half lying, and standing while touching objects in front.
Retrieved human body models $h_i^C$ are associated to their \textit{contact object} $o_i^C$ (the apex $C$ stands for ``contact''). Eventually, 3D scenes are derived by fitting the object to the layouts. For visualization purposes only, we optionally populate the scenes with human body models.

For human-aware 3D scene synthesis, most solutions train neural networks on well-aligned 3D scenes and human motion sequences. This is impractical, because the vast number of possible trajectories results in time-consuming and memory-intensive processing. Instead, we consider \textit{local scenes}, i.e., scene elements surrounding those objects with which a human is most likely to interact. We also observed that furniture objects with similar functionality are usually spatially arranged together. Therefore, we define functional object groups $\Omega$, e.g., \textit{bed}-and-\textit{nightstand}, \textit{table}-and-\textit{chair}, \textit{coffee table}-and-\textit{sofa}. The detailed definition of $\Omega$ can be found in the \underline{supplementary material}. By categorizing objects into disjoint groups, our method enables modeling both spatial and functional co-occurrence.

More in detail, our strategy for human-aware 3D scene refinement encourages appropriate human-object contact and discourages human-object surface interpenetration. The overall process is formalized in Algorithm: given a 3D scene with $n$ objects to be optimized, we consider the $q\leq n$ humans $h_i^C$, each in contact with its own \textit{contact object}  $o_i^C$ as defined in Sec. \ref{subsec:assembly}. For all possible human-object interactions we check whether humans intersect with objects which are different from their own \textit{contact object}, since intersections between humans and objects have to be discouraged. Specifically, objects of the same category $c_i$ are moved far apart from the human, while overlapping objects within the same functional object group $\Omega$ are adjusted. Finally, if a human intersects an object not in the same functional object group $\Omega$ as its \textit{contact object}, the former is discarded as misplaced. We report here the pseudo-code for the optimization process of the spatial critic, in Algorithm~\ref{alg:optimization}.

\begin{center}
\begin{minipage}{0.9\linewidth}
\begin{algorithm}[H]
\caption{Spatial Critic Optimization}
\label{alg:optimization}
\begin{algorithmic}[1]
\State \textbf{Input}: scene $S=\{o_j\}_{j=1}^n$, object categories $\{c_j\}_{j=1}^n$,
\Statex \hspace{\algorithmicindent} contact humans $H^C=\{h_i^C\}_{i=1}^q$, contact objects $\{o_i^C\}_{i=1}^q$,
\Statex \hspace{\algorithmicindent} functional object groups $\Omega$, threshold $\beta$.

\For{$h_i^C \in H^C$}
    \For{$j=1,\ldots,n$}
        \If{$h_i^C \cap o_j \neq \varnothing$ \textbf{and} $o_i^C \neq o_j$}
            \If{$c_j = c_i$}
                \State move $o_j$ away from $h_i^C$
            \ElsIf{$c_j \neq c_i$ \textbf{and} $(o_i^C,o_j)\in\Omega$}
                \If{intersection area $\geq \beta$}
                    \State move $o_j$ away from $h_i^C$
                \EndIf
            \ElsIf{$c_j \neq c_i$ \textbf{and} $(o_i^C,o_j)\notin\Omega$}
                \State remove $o_j$
            \EndIf
        \EndIf
    \EndFor
\EndFor
\end{algorithmic}
\end{algorithm}
\end{minipage}
\end{center}

\begin{figure*}[t]
  \centering
   \includegraphics[width=.96\linewidth]{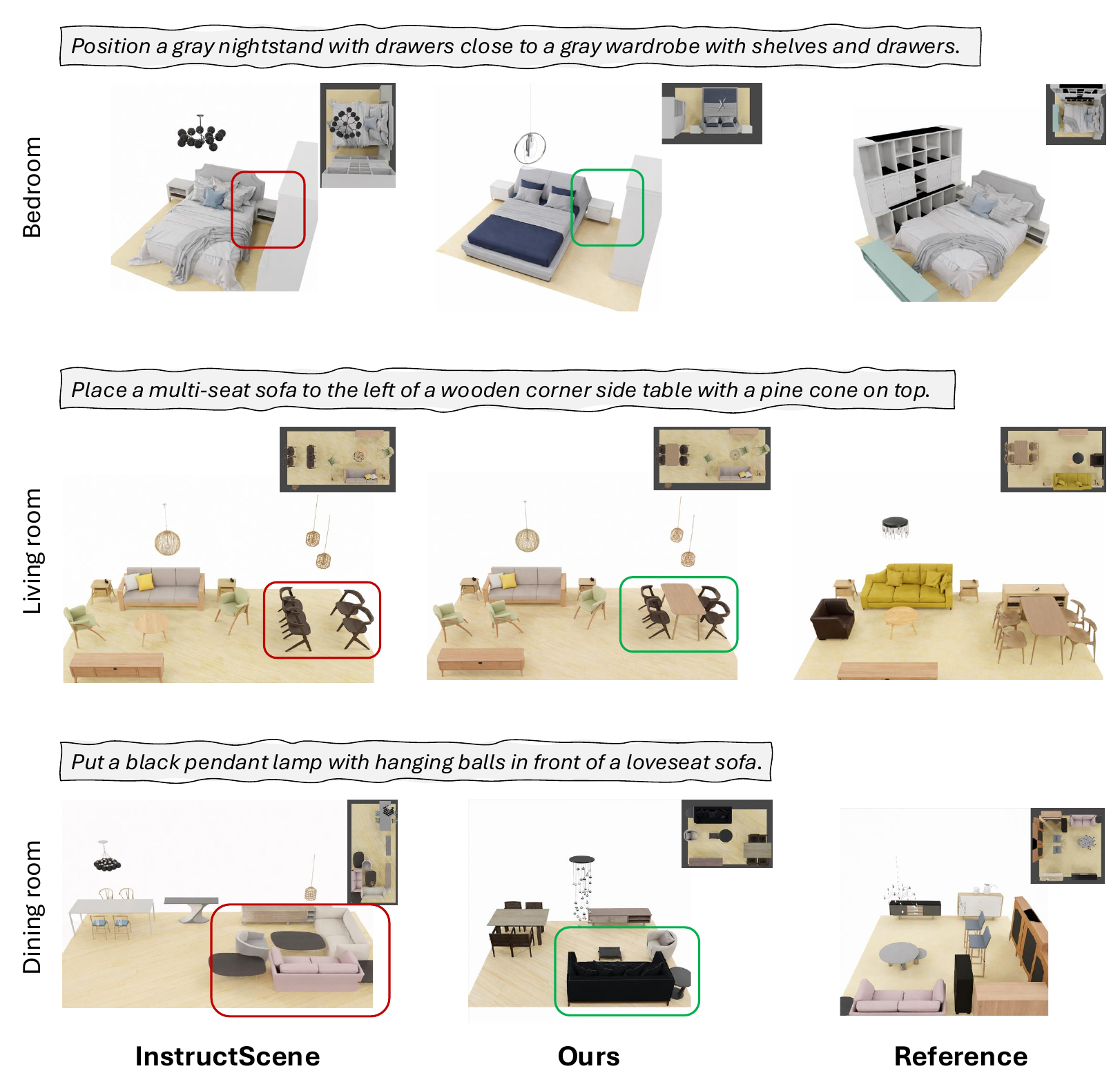}
   \caption{Qualitative results. Our \modelname{} is able to recover from inaccurate layouts (e.g., nightstand tightly placed in between the bed and wardrobe in the first example or overlapping chairs in the second one) which are instead present in InstructScene~\cite{lin2024instructscene}, our baseline. As a reference, we show the ground-truth arrangement. Best viewed in color.}
   \label{fig:result}
\end{figure*}

\section{Experiments}
\label{sec:exp}

\subsection{Setup and Protocol}

\noindent\textbf{Dataset.} The 3D-FRONT dataset \cite{fu20213d, fu20213d2} is a large-scale repository of indoor synthetic scenes with professionally designed furniture layouts and with textured 3D models available in different styles. The dataset contains $18k$ different rooms, and the evaluation protocols focus on \emph{bedroom}, \emph{dining room} and \emph{livingroom} scenes, which amounts to 4041, 900 and 813 synthetic rooms respectively. The number of object classes available varies with the type of scene: bedrooms typically include 3 to 12 instances from 21 object categories, while for living rooms and dining rooms there are from 3 to 21 instances from 24 categories. Following InstructScene~\cite{lin2024instructscene}, the training and testing data are split as follows: 3879/162 for the bedroom scenes, 621/192 for the living room scenes, and 723/177 for the dining rooms scenes. Each scene is paired with a text caption consisting of triplets. We adopt the same captioning and sampling strategies as InstructScene~\cite{lin2024instructscene} for a fair comparison. The objects are automatically captioned using a combination of BLIP~\cite{blip2022} and ChatGPT~\cite{ouyang2022training}, while their spatial relationships are extracted using a set of hard-coded rules resulting in 11 possible relationships (e.g., left of, right of, above, below). Then, a text prompt can be formed by sampling one or two triplets \(\langle \text{subject}, \text{predicate}, \text{object} \rangle\). For inference, we utilize the same text prompts as InstructScene.

\begin{table}[tb]
  \caption{Comparisons with text-to-3D scene generation baselines on the 3D-FRONT dataset using commonly reported metrics. Best scores are \textbf{bold}. We provide standard deviation values by subscripts. $^\dagger$ denotes official results since the code is not open-sourced.
  \label{tab:main_results}}
  \centering
  \small
  \begin{tabular}{@{}cl|ccccc@{}}
    \hline
    \multicolumn{2}{c|}{\textbf{Method}} & $\uparrow$ iRecall$_\%$ & $\downarrow$ FID & $\downarrow$ FID-C & $\downarrow$ KID$_{\times\texttt{1e-3}}$ & SCA$_\%$\\
    \hline\hline
    \multirow{5}{*}{\rotatebox{90}{Bedroom}} & ATISS \cite{paschalidou2021atiss} & $48.13_{\pm2.50}$ & $119.73_{\pm1.55}$ & $6.95_{\pm0.06}$ & $0.39_{\pm0.02}$ & $59.17_{\pm1.39}$ \\
    & DiffuScene \cite{tang2024diffuscene} & $56.43_{\pm2.07}$ & $123.09_{\pm0.79}$ & $7.13_{\pm0.16}$ & $0.39_{\pm0.01}$ & $60.49_{\pm2.96}$ \\
    & InstructScene \cite{lin2024instructscene} & $73.64_{\pm1.37}$ & $114.78_{\pm1.19}$ & $6.65_{\pm0.18}$ & $\mathbf{0.32_{\pm0.03}}$ & $56.02_{\pm1.43}$ \\
    & FreeScene$^\dagger$\cite{bai2025freescene} & 73.69 & \textbf{111.21} & \textbf{6.43} & 0.35 & 54.94 \\
    & \modelname{} (ours) & $\mathbf{76.80_{\pm1.25}}$ & $\mathbf{111.67_{\pm0.69}}$ & $\mathbf{6.39_{\pm0.31}}$ & $0.79_{\pm0.06}$ & $\mathbf{52.11_{\pm0.84}}$ \\
    \hline
    \multirow{5}{*}{\rotatebox{90}{Living room}} & ATISS \cite{paschalidou2021atiss} & $29.50_{\pm3.67}$ & $117.67_{\pm2.32}$ & $6.08_{\pm0.13}$ & $17.60_{\pm2.65}$ & $69.38_{\pm3.38}$ \\
    & DiffuScene \cite{tang2024diffuscene} & $31.15_{\pm2.49}$ & $122.20_{\pm1.09}$ & $6.10{\pm0.11}$ & $16.49_{\pm1.24}$ & $72.92_{\pm1.29}$ \\
    & InstructScene \cite{lin2024instructscene} & $56.81_{\pm2.85}$ & $110.39_{\pm0.78}$ & $\mathbf{5.37_{\pm0.07}}$ & $8.16_{\pm0.56}$ & $65.42_{\pm2.52}$ \\
    & FreeScene$^\dagger$\cite{bai2025freescene} & \textbf{58.16} & 110.55 & 5.83 & 7.95 & \textbf{55.24} \\
    & \modelname{} (ours) & $57.24_{\pm0.47}$ & $\mathbf{105.01}_{\pm0.63}$ & $\mathbf{5.24_{\pm0.24}}$ & $\mathbf{6.78_{\pm0.32}}$ & $65.72_{\pm0.73}$ \\
    \hline
    \multirow{5}{*}{\rotatebox{90}{Dining room}} & ATISS \cite{paschalidou2021atiss} & $37.58_{\pm1.99}$ & $137.10_{\pm0.34}$ & $8.49_{\pm0.23}$ & $23.60_{\pm2.52}$ & $67.61_{\pm3.23}$ \\
    & DiffuScene \cite{tang2024diffuscene} & $37.87_{\pm2.76}$ & $145.48_{\pm1.36}$ & $8.63_{\pm0.31}$ & $24.08_{\pm1.90}$ & $70.57_{\pm2.14}$ \\
    & InstructScene \cite{lin2024instructscene} & $61.23_{\pm1.67}$ & $129.76_{1.61}$ & $7.67_{\pm0.18}$ & $13.24_{\pm1.79}$ & $64.20_{\pm1.90}$ \\
    & FreeScene$^\dagger$\cite{bai2025freescene} & 63.39 & 127.28 & 8.01 & 14.83 & \textbf{56.82} \\
    & \modelname{} (ours) & $\mathbf{65.02_{\pm0.28}}$ & $\mathbf{123.07_{\pm0.64}}$ & $\mathbf{6.84_{\pm0.36}}$ & $\mathbf{10.99_{\pm0.85}}$ &  $60.86_{\pm0.95}$ \\
  \hline
  \end{tabular}
\end{table}

\noindent\textbf{Baseline.} To assess our model, we compare its performance against ATISS \cite{paschalidou2021atiss}, an auto-regressive model that sequentially generates unordered object sets; DiffuScene \cite{tang2024diffuscene}, which treats scene attributes as continuous 2D matrices with a Gaussian diffusion model; InstructScene \cite{lin2024instructscene}, which learns common sense of indoor scenes using graph diffusion models; and FreeScene \cite{bai2025freescene}, which employs a mixed graph diffusion process with constrained sampling. 

\noindent\textbf{Metric.} For each of the three types of scenes, bedroom, living room and dining room, we report the same evaluation metrics used in InstructScene \cite{lin2024instructscene}. These are instruction recall (iRecall), Fréchet Inception Distance (FID), FID-C which computes FID by CLIP features, Kernel Inception Distance (KID), and scene classification accuracy (SCA). To reflect the quality of constructed scene graphs, iRecall quantifies the fraction of \(\langle \text{subject}, \text{predicate}, \text{object} \rangle\) triplets present in the prompt that are also present in the synthesized scenes, i.e., the higher the metric the more the information provided by the prompt is preserved in the scene. FID, FID-C and KID scores, computed using the clean-fid library \cite{parmar2022aliased}, measure the similarity between the learned and real distributions of top-down renderings (at $256^2$ resolution), with lower scores indicating greater similarity. Finally, SCA is obtained by fine-tuning an AlexNet \cite{krizhevsky2012imagenet}, pretrained on ImageNet \cite{deng2009imagenet}, to discriminate between ground truth and synthesized scenes, and hence the closest SCA is to $50\%$ accuracy, the better the synthesized scenes are. 

\begin{figure}[tb]
  \centering
   \includegraphics[width=\linewidth]{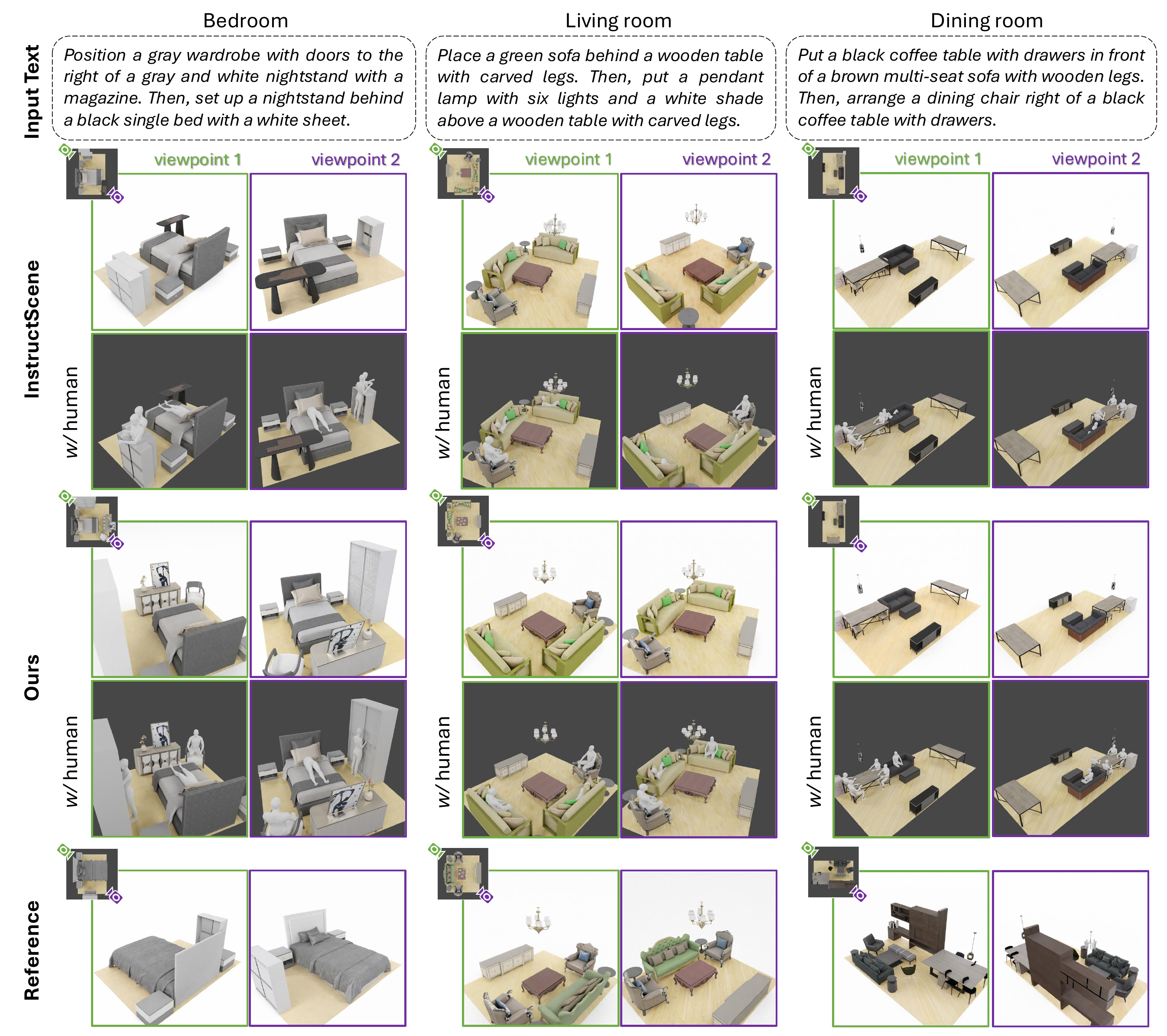}
   \caption{Additional 3D scene results rendered from diverse perspectives (populated with 3D human bodies). InstructScene tend to produce repeating wardrobes (left), overlapping coffee table and sofa (middle), and overlapped dining chairs (right). Instead, the 3D scenes synthesized by our \modelname{} shows more realistic arrangements with less collisions. Best viewed in color.}
   \label{fig:res_add_supp}
\end{figure}

\begin{table}[t]
\caption{Evaluation on physical metrics~\cite{yang2024physcene}. The best results are \textbf{bold}.}
\label{exp:physical}
\centering
\small
\setlength{\tabcolsep}{4.5pt}
\begin{tabular}{c|cc|cc|cc}
\hline
\multirow{2}{*}{\textbf{Metric}} 
& \multicolumn{2}{c|}{\textbf{Bedroom}} 
& \multicolumn{2}{c|}{\textbf{Living room}} 
& \multicolumn{2}{c}{\textbf{Dining room}} \\
& InstructScene~\cite{lin2024instructscene} & Ours
& InstructScene~\cite{lin2024instructscene} & Ours
& InstructScene~\cite{lin2024instructscene} & Ours \\
\hline\hline
$Col_{obj}$   ($\downarrow$) & 0.21 & \textbf{0.17} & 0.23 & \textbf{0.19} & 0.26 & \textbf{0.18} \\
$Col_{scene}$ ($\downarrow$) & 0.39 & \textbf{0.31} & 0.55 & \textbf{0.51} & 0.61 & \textbf{0.44} \\
\hline
\end{tabular}
\end{table}

\begin{table*}[tb]
\caption{Ablation studies on dining rooms. The best results are \textbf{bold}.
\label{exp:ablation}}
\centering
\small
\begin{tabular}{c|cccccc}
\hline
Settings & $\uparrow$ iRecall$_\%$ & $\downarrow$ FID & $\downarrow$ FID-C & $\downarrow$ KID & SCA$_\%$ \\
\hline\hline
Full model (ours) & \textbf{65.06} & \textbf{122.54} & \textbf{6.72} & \textbf{10.57} &  61.93 \\
w/o \textit{Semantic Critic} & 64.31 & 130.14 &  7.88 & 12.61 & \textbf{60.22} \\
w/o \textit{Spatial Critic} & \textbf{65.06} & 124.45 & 6.94 & 11.55 & 64.20 \\
Baseline & 60.59 & 130.10 &  7.81 &  13.65 & 66.48 \\
\hline
\end{tabular}
\end{table*}

\begin{figure}[tb]
  \centering
   \includegraphics[width=.85\linewidth]{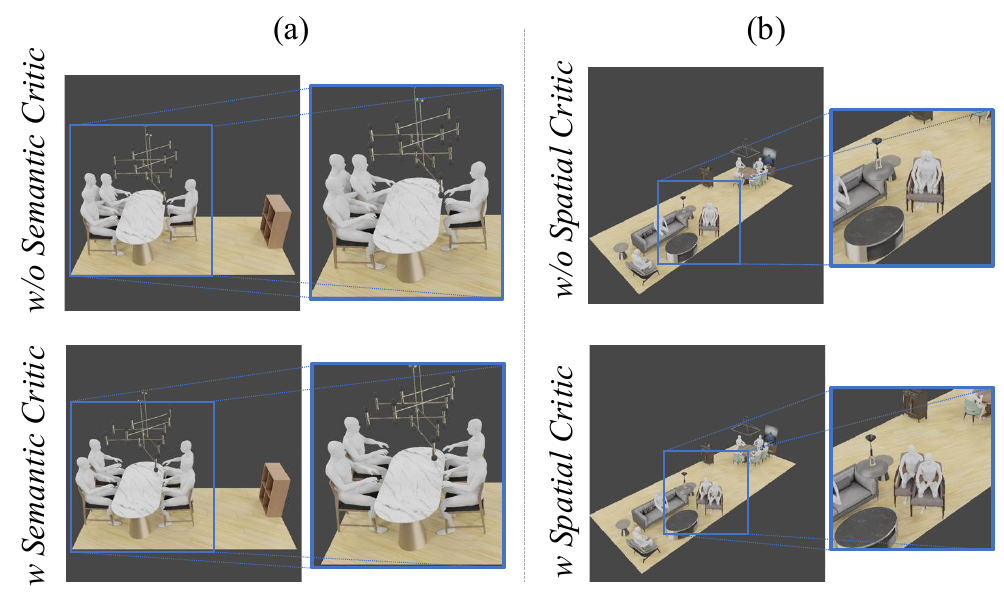}
   \caption{(a) Applying semantic critic results in more structured layouts than naive layout estimation. (b) The spatial critic introduces benefits for local spatial arrangements. For a detailed view, the content highlighted in the blue box is shown in a zoomed-in format.}
   \label{fig:result_abla_reas}
\end{figure}

Additional setup details are provided in the \underline{supplementary material}.

\subsection{3D Scene Generation Results}

Table~\ref{tab:main_results} shows how our method increases performance compared to previous methods for text-to-3D scene synthesis. Compared to the baseline method InstructScene \cite{lin2024instructscene}, the proposed method consistently shows superior iRecalls, FID, FID-C and the best SCAs, i.e., closest to $50\%$, over the three scene types, confirming the overall efficacy of our approach. While our method generally achieves good results, it exhibits slight KID increases on the ``Bedroom'' scenes, which usually contain less objects, but outperforms the competitors on the other challenging scene types. For the physical metrics, our \modelname{} outperforms the baseline across diverse room types.

Additionally, we employ the metrics introduced bycPhyScene~\cite{yang2024physcene}, including the percentage of objects that collide with other objects in the generated scene ($Col_{obj}$) and the proportion of generated scenes containing at least one object collision ($Col_{scene}$). Table~\ref{exp:physical} reports the corresponding physical plausibility results. Compared with InstructScene~\cite{lin2024instructscene}, our method consistently achieves lower collision rates on both metrics across bedroom, living room, and dining room scenes. These results demonstrate that our interaction-driven spatial refinement effectively improves the physical validity of generated 3D scenes.

To further showcase the performance of our method with respect to the baseline provided by InstructScene~\cite{lin2024instructscene}, we showcase in Fig.~\ref{fig:result} one example of synthesized room for each type of scene. Thanks to the intermediate graphs, both methods can reason about scene content based on the input implicit prompts. As a result, one or two objects and relationships are mentioned in the prompt, while the generated scenes typically include more objects and relationships. This inconsistency, often perceived as diversity in generative modeling, enables greater flexibility and creativity in synthesizing 3D indoor scenes. It is observed that InstructScene fails to synthesize functional 3D scenes, such as overlapped objects (in bedroom and living room), missing functional objects (dining table in living room), and overly-generated objects (three coffee tables in dining room). In contrast, the scenes synthesize by our method exhibit higher quality, for instance, a dining table is surrounded by dining chairs, which belong to the same functional object group, supporting human action \textit{sitting}. Furthermore, our method achieves more similar results to the reference, ground-truth scenes from 3D-FRONT. Overall this suggests our model can provide more ``physcially-realistic'' scenes.

As shown in Fig.~\ref{fig:res_add_supp}, in addition to the top-down rendering of each scene, we provide zoomed-in additional renderings (at $1024^2$ resolution) from viewpoints at opposite corners of the rooms, i.e., a camera at top-left and one at the bottom-right corner of the scenes, looking down from a high vantage point. To further highlight how our scenes are more compatible with human-object interactions, we also show the same synthesized 3D scenes populated with the retrieved 3D human poses. Compared to InstructScene, our \modelname{} can generate visually appealing and functionally coherent scenes which supports appropriate human-object contacts, enabling the practical usability of the generated environment.

\subsection{Ablations}

We ablate two core components of our framework: Semantic Critic and Spatial Critic. Table~\ref{exp:ablation} compares our full \modelname{} model with variants where each module is removed. It can be observed that introducing these components solely can not lead to significant improvements against the baseline~\cite{lin2024instructscene}. The best overall results are achieved when both critics are enabled, demonstrating the complementary benefits of semantic consistency checking and physically grounded interaction validation. 

Fig.~\ref{fig:result_abla_reas} shows two groups of examples: (a) without and with \textit{Semantic Critic} and (b) without and with \textit{Spatial Critic}. We observe that introducing these components solely can not lead to significant improvements against the baseline. While the Semantic Critic alleviates the inaccurate one-shot layout estimation, it may struggle with fine-grained spatial arrangements. The Spatial Critic improves object-level functionality, but lacks sufficient global awareness. In contrast, our full model performs superior to ablated versions with certain components removed, verifying the effectiveness of the presented components.

\begin{table}[tb]
\caption{Quantitative evaluation of our and state-of-the-art indoor Text-to-3D Scene Synthesis models on zero-shot application to downstream tasks - Stylization, Re-arrangement, Completion, and Unconditional generation. We highlight the \textbf{best} and \underline{second best} performance.
\label{tab:zeroshot}}
\centering
\small
\begin{tabular}{@{}cl|c|cc|cc|c@{}}
\hline
\multicolumn{2}{c|}{\multirow{2}{*}{\textbf{Applications}}} & \textbf{Stylization} & \multicolumn{2}{c|}{\textbf{Re-arrangement}} & \multicolumn{2}{c|}{\textbf{Completion}} & \textbf{Uncond.}\\
& & $\downarrow$ FID & $\uparrow$ iRecall$_\%$ & $\downarrow$ FID & $\uparrow$ iRecall$_\%$ & $\downarrow$ FID & $\downarrow$ FID \\ 
\hline\hline
\multirow{5}{*}{\rotatebox{90}{Bedroom}} & ATISS \cite{paschalidou2021atiss} & 123.91 & 61.22 & 107.67 & 64.90 & 89.77 & 134.51 \\
& DiffuScene \cite{tang2024diffuscene} & 127.35 & 68.57 & 106.15 & 48.57 & 96.28 & 135.46 \\
& InstructScene \cite{lin2024instructscene} & \underline{122.73} & \underline{79.59} & \underline{105.27} & \underline{69.80} & 82.98 & \underline{124.97} \\
& FreeScene \cite{bai2025freescene} & - & 77.49 & 105.71 & 69.25 & \textbf{81.83} & \textbf{123.42} \\
& \modelname{} (ours) & \textbf{120.46} & \textbf{84.90} & \textbf{101.37} & \textbf{69.84} & \underline{82.05} & 126.02 \\
\hline
\multirow{5}{*}{\rotatebox{90}{Living room}} & ATISS \cite{paschalidou2021atiss} & 110.85 & 31.97 & 117.97 & 43.20 & 106.48 & 129.23 \\
& DiffuScene \cite{tang2024diffuscene} & 112.80 & 41.50 & 115.30 & 19.73 & 95.94 & 129.75 \\
& InstructScene \cite{lin2024instructscene} & \underline{109.39} & \underline{56.12} & \underline{106.03} & \underline{46.94} & 92.52 & 117.62 \\
& FreeScene \cite{bai2025freescene} & - & 54.82 & 106.33 & \textbf{50.57} & \underline{91.01} & \underline{116.87} \\
& \modelname{} (ours) & \textbf{107.47} & \textbf{56.94} & \textbf{104.98} & 45.58 & \textbf{90.64} & \textbf{115.11} \\
\hline
\multirow{5}{*}{\rotatebox{90}{Dining room}} & ATISS \cite{paschalidou2021atiss} & 131.14 & 36.06 & 134.54 & 57.99 & 122.44 & 147.52 \\
& DiffuScene \cite{tang2024diffuscene} & 135.20 & 46.84 & 133.73 & 32.34 & 115.08 & 150.81 \\
& InstructScene \cite{lin2024instructscene} & \underline{128.78} & 62.08 & 125.07 & \underline{60.59} & \underline{107.86} & 137.52 \\
& FreeScene \cite{bai2025freescene} & - & \underline{63.02} & \underline{124.69} & \textbf{60.67} & \textbf{105.71} & \textbf{132.22} \\
& \modelname{} (ours) & \textbf{127.63} & \textbf{65.54} & \textbf{122.32} & 56.77 & 111.08 & \underline{136.85}\\
\hline
\end{tabular}
\end{table}

\subsection{Zero-shot Applications}

The refined scene graph generated in the Reasoning stage can also be used for downstream tasks, without further fine-tuning. To highlight this, we follow \cite{lin2024instructscene} and apply our pre-trained model to four zero-shot tasks - stylization, re-arrangement, completion,and unconditional generation. 
This downstream test did not require further training.
The first three tasks generate scenes conditioning them also on some pre-existing features, in addition to the text prompt $X$.
Stylization $p_\phi(f|\lambda,c,a,t,s,r)$ predicts the texture and appearance features $f$ given the layout $(t,s,r)$, the human actions $a$ and the object categories $c$. Similarly, rearrangement $p_{\phi,\theta}(t,s,r|\lambda,c,f,a)$ predicts the layout given all other scene's features, and scene completion introduces new objects to the scene while preserving the existing ones and their features. Unconditional generation conditions a scene from null textual features $\lambda$, producing a scene dependent exclusively on the distributions learned from the dataset. 

For all four applications, we report the quantitative evaluation in Table \ref{tab:zeroshot}, comparing it against FreeScene \cite{bai2025freescene}, InstructScene \cite{lin2024instructscene}, ATISS \cite{paschalidou2021atiss} and DiffuScene \cite{tang2024diffuscene}. This evaluation shows how our method achieves the best performances in all tasks on the Bedroom scenes. Additionally, in the living room scenes - which are usually richer in objects and relationships - our method has the best FIDs on stylization, completion and unconditional generation tasks. Qualitative comparisons on these zero-shot applications are provided in the \underline{supplementary material}.

\section{Conclusion}
\label{sec:con}

We presented \modelname{}, a compositional framework for generating 3D indoor scenes from implicit text descriptions. Instead of treating text-to-scene synthesis purely as a layout prediction problem, our method first reasons over scene graphs and then refines the generated layouts with interaction-driven semantic and spatial critics. Experiments on the 3D-FRONT dataset demonstrate that \modelname{} achieves competitive or state-of-the-art performance across standard scene generation metrics, while also improving physical plausibility by reducing object collisions. These results suggest that incorporating lightweight human-object interaction priors is an effective way to generate indoor scenes that are not only visually realistic, but also more functionally coherent.

\noindent\textbf{Limitations.} The presented work focuses primarily on object layout by leveraging complementary critics that enable self-correction, improving both semantic plausibility and spatial arrangement. However, 3D assembly is performed via a shape retrieval paradigm, which inherently restricts object diversity and limits the ability to generate novel object geometries. Additionally, richer interactions such as accessibility are not explicitly optimized.

\noindent\textbf{Future work.} Our method extends to richer compositional representations for scene generation. Future work includes adapting the framework to other indoor environments, such as factories or hospitals, by generative object generation and motion-aware layout refinement. In addition, incorporating a Mixture-of-Experts (MoE) \cite{bai2023qwen} architecture could further improve the model by enabling more structured and hierarchical expert specialization, leading to better disentanglement of semantics, geometry, and physical constraints. This may enhance scalability to larger and more diverse 3D scenes.

\clearpage

\bibliography{egbib}

\clearpage

\appendix

\begin{center}
    {\Large \textbf{Supplementary Material}}
\end{center}

This supplementary material provides additional implementation details, qualitative results and analysis, as well as zero-shot applications to support the main paper and facilitate reproducibility.

\section{Implementation Details}

\begin{figure}[h]
  \centering
   \includegraphics[width=\linewidth]{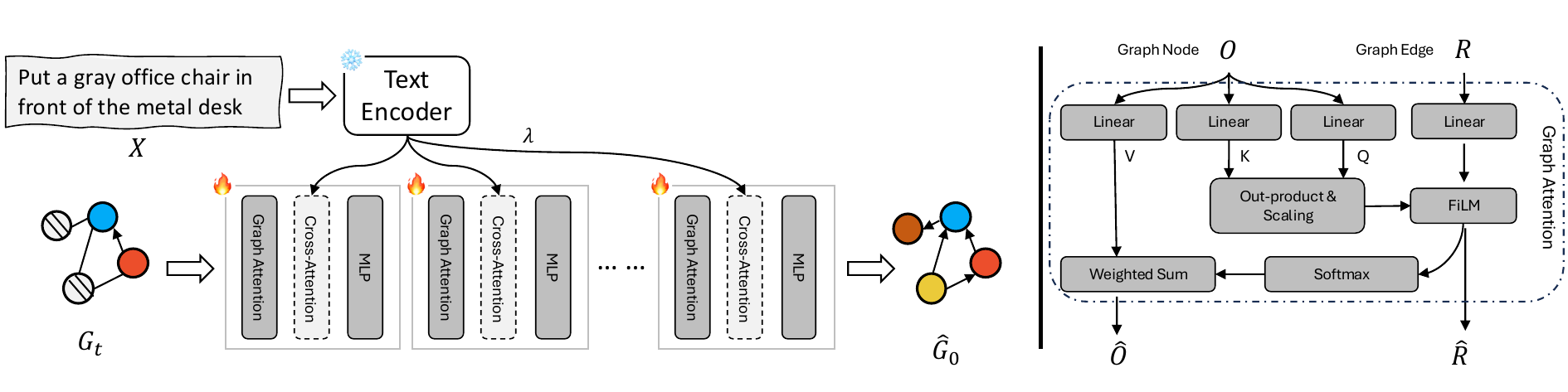}
   \caption{\textit{Left}: The architecture of graph transformer which serves as a denoiser network. \textit{Right}: The detail of graph attention layer.}
   \label{fig:transformer}
\end{figure}

\noindent\textbf{Graph diffusion.} The proposed pipeline consists of two denoising diffusion models, $\phi$ and $\theta$, which are trained separately and used respectively in the Semantic Reasoning and Spatial Assembly stages. Following the baseline InstructScene, the denoiser network is implemented as a 5-layer and 8-head transformer with an attention dimension of 512 and a dropout rate of 0.1. As shown in Fig.~\ref{fig:transformer}, the graph transformer adopts a stacked-block design, where each block integrates graph attention and an MLP. Furthermore, for the text-to-graph diffusion model $\phi$, cross-attention layers are inserted after the graph attention layers to incorporate textual guidance. Both models are trained on a single GPU with a batch size of 128, using as optimizer AdamW with a learning rate of $1\mathrm{e}{-4}$, and exponentially moving average (EMA) technique with a decay factor of 0.9999.

\noindent\textbf{Semantic critic.} We parse the input text using a dependency parser to identify object categories and object relations explicitly mentioned in the prompt. During graph diffusion, we enforce that these anchors appear in the generated scene graph. If the sampled graph does not contain the required anchors, it is rejected and re-sampled until the anchor constraints are satisfied. At the same time, we assess functional plausibility through human-object interactions. Each graph node $o_i=\{c_i,f_i,a_i\}$ includes an object category $c_i$, visual features $f_i$, and potential human actions $a_i$. The action labels $a_i$ are predicted using Llama-2 given the room type and the object list. Compared with earlier human-to-scene generation methods that rely on human sequences as inputs, our approach infers potential interactions directly from text, enabling functional reasoning without requiring complex human trajectory annotations.

\noindent\textbf{Spatial critic.} In the main paper, we introduced functional object groups $\Omega$, which aggregate objects with similar
functionality that are often found in close proximity, potentially leading to more human-object interactions. For example, \textit{dining chairs} are usually close to \textit{dining tables}, and a human \textit{sitting} on the former will likely be also in contact with the latter. As shown in Table~\ref{tab:group}, we define groups based on functional co-occurrence rules, \eg reception-oriented (\eg \textit{coffee table-and-sofa}). Such definitions are used to include soft constraints during human-scene optimization. The threshold $\beta$ is set to $30\%$.

\begin{figure}[tb]
  \centering
   \includegraphics[width=.96\linewidth]{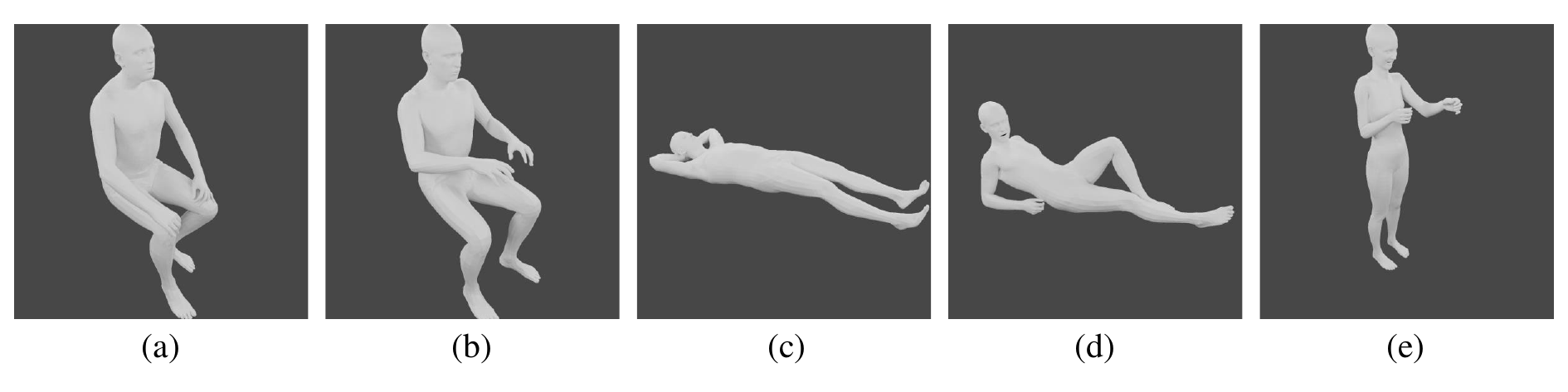}
   \caption{3D humans collection, including their actions (a) \textit{sitting}, (b) \textit{sitting and touching}, (c) \textit{lying}, (d) \textit{half lying}, and (e) \textit{standing while touching}.}
   \label{fig:res_huma}
\end{figure}

\noindent\textbf{Shape retrieval.} Object retrieval is done by filtering the 3D-FUTURE dataset based on the query category $c$ and filtering models by feature cosine similarities, i.e., prioritizing those that more closely match the geometry and appearance of the query feature $f$. Given the top-K most similar textured 3D models, we select the one closest to the predicted query size $s$. Instead, human meshes are retrieved from a collection of SMPL-X models of static poses from RenderPeople scans (Fig.~\ref{fig:res_huma}). Each \textit{contact object} is assigned one of five \textit{contact humans} according to the human action $a$ and the object category $c$: for example, \textit{sitting} in a leisure context, \eg on a sofa, will match to Fig.~\ref{fig:res_huma}a, while sitting at a \textit{desk} or \textit{table} will match to Fig.~\ref{fig:res_huma}b. The contact humans share the same poses $(t,r)$ of the matched objects, with the exception of Fig.~\ref{fig:res_huma}e, usually associated with objects like \textit{wardrobe} and \textit{cabinet}. In this case, the human is rotated by  $180^{\circ}$ to place it facing towards the object it is touching (e.g., open or close the door).

\begin{table}[ht]
    \centering
    \caption{Object functional groups.}
    \begin{tabular}{c|p{6cm}}
        \hline
        Function & Groups \\
        \hline\hline
        relax & double bed-and-night stand; single bed-and-night stand; lazy sofa-and-table; L-shape sofa-and-table;\\
        \hline
        dining & dining table-and-dining chair; dining table-and-wine cabinet; dining table-and-shelf; desk-and-chair; \\
        \hline
        reception & coffee table-and-sofa; coffee table-and-lounge chair; coffee table-and-multi seat sofa; coffee table-and-armchair; \\
        \hline
    \end{tabular}
    \label{tab:group}
\end{table}

\section{Additional Results}

\begin{figure*}[h]
  \centering
   \includegraphics[width=\linewidth]{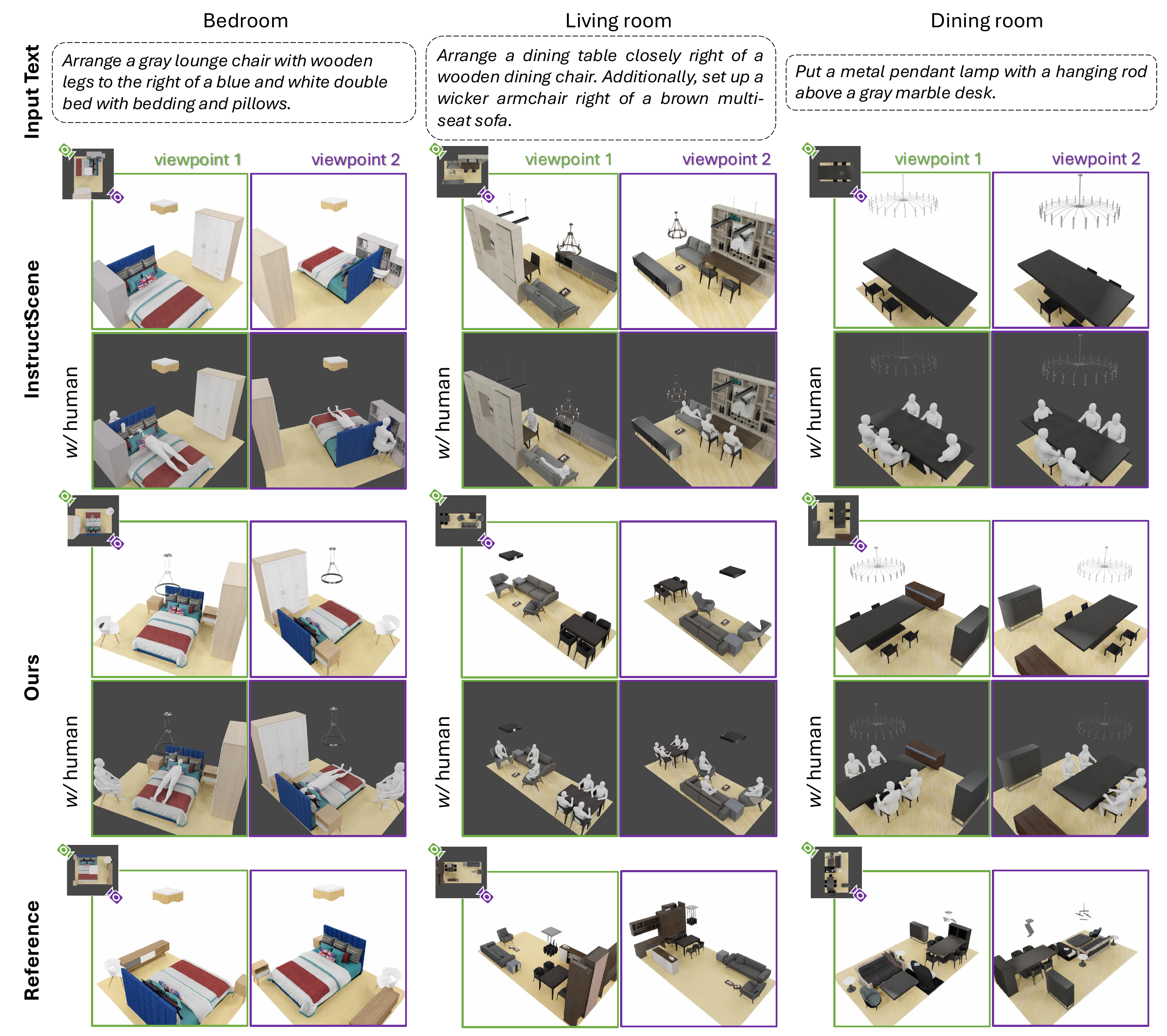}
   \caption{Additional scene results rendered from diverse perspectives (populated with 3D human bodies). Compared to InstructScene which results in overlapped objects, our method shows more realistic arrangements with less collisions.}
   \label{fig:res_add}
\end{figure*}

\noindent\textbf{Qualitative results.} To provide more context for the results of the paper and support the claims about the quality of our synthetic scenes, we expand on the qualitative results provided in the main paper. As shown in Fig.~\ref{fig:res_add}, \modelname{} can generate visually appealing and functionally coherent scenes which supports appropriate human-object contacts, enabling the practical usability of the generated scene.

\begin{figure*}[t]
  \centering
   \includegraphics[width=.86\linewidth]{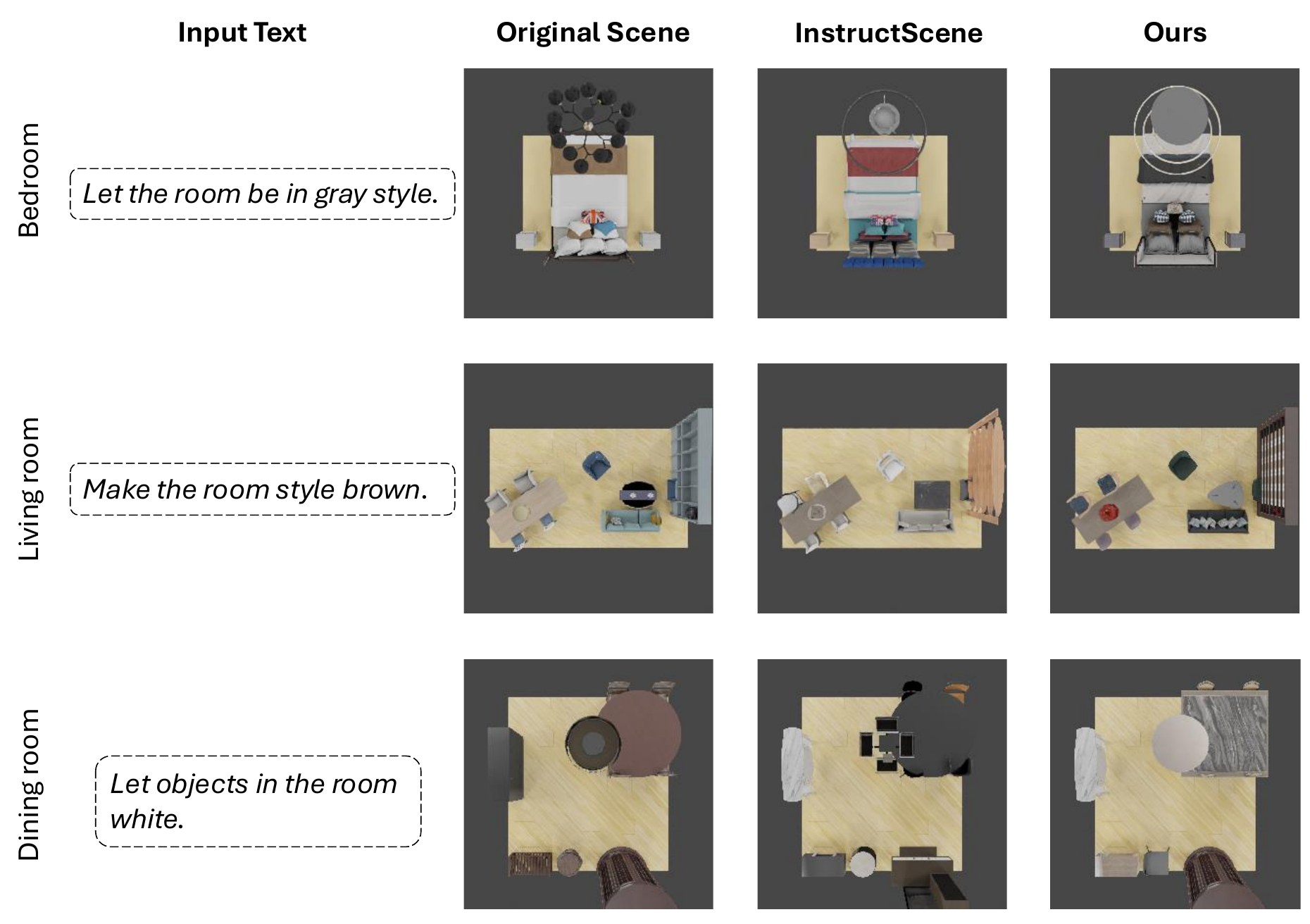}
   \caption{Stylization results. Given an original scene and a text description, the object features can be changed while keeping the other scene elements unchanged. Compared against InstructScene, our \modelname{} exhibits higher consistency with the input text.}
   \label{fig:res_zero_styl}
\end{figure*}

\begin{figure*}[t]
  \centering
   \includegraphics[width=.86\linewidth]{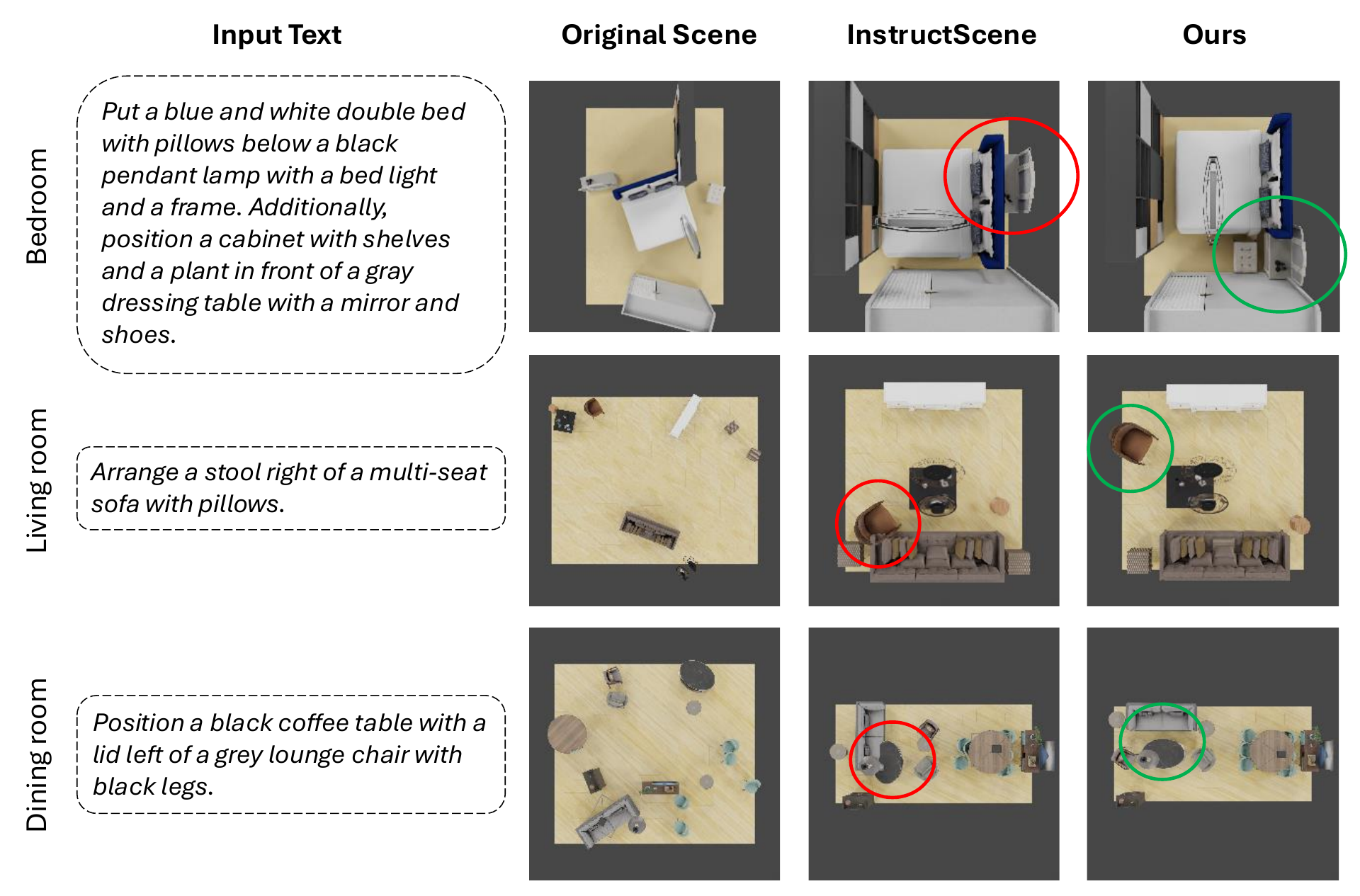}
   \caption{Re-arrangement results. The object layouts can be re-generated conditioning on input text while keeping the other scene elements unchanged. As shown in green/red circles, our \modelname{} results in more realistic spatial arrangements than InstructScene.}
   \label{fig:res_zero_rear}
\end{figure*}

\begin{figure*}[t]
  \centering
   \includegraphics[width=.86\linewidth]{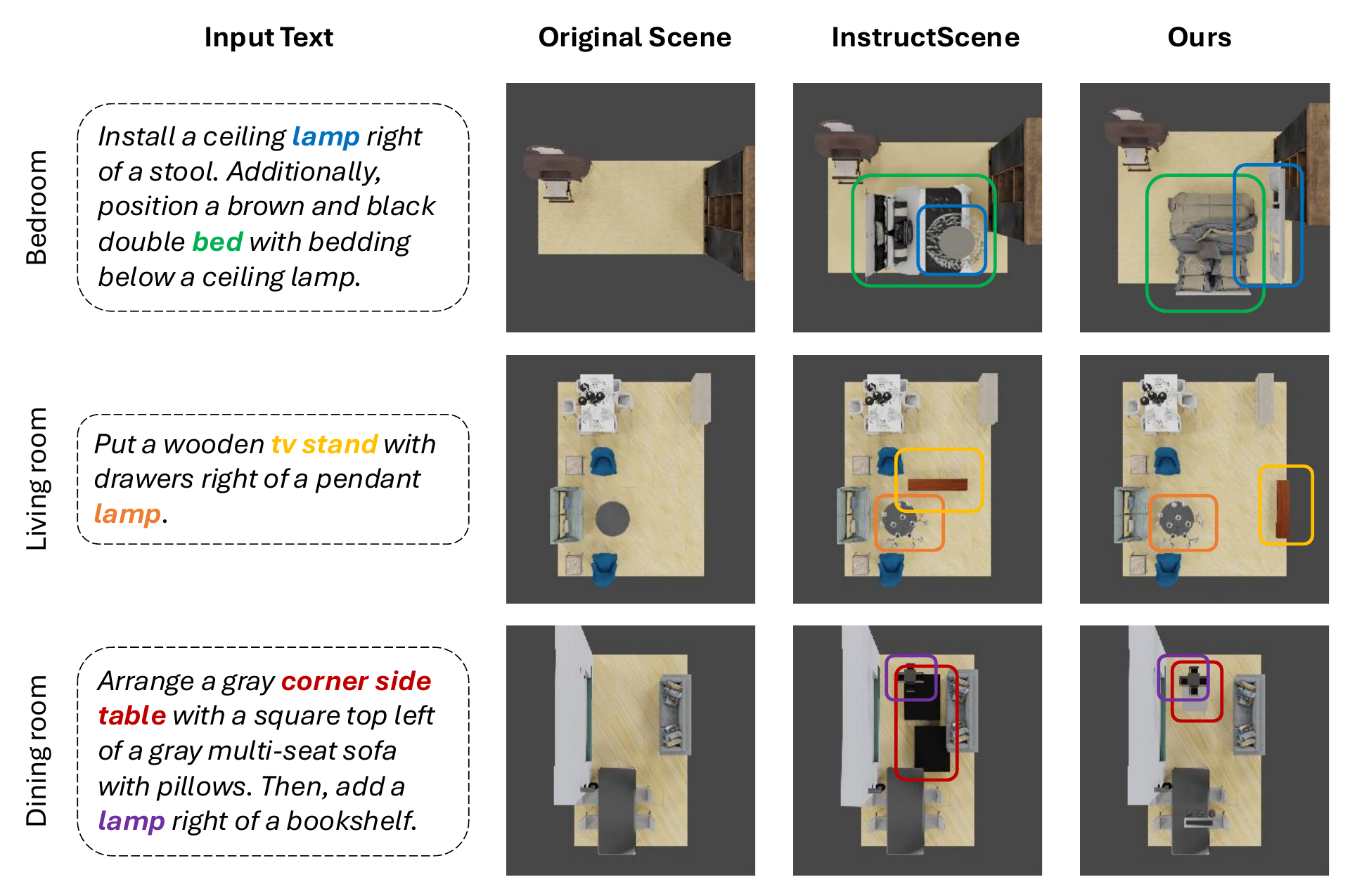}
   \caption{Completion results. New objects can be incorporated into the existing 3D scene conditioned on input text. Our \modelname{} generates more practical object poses, \eg, the bed and the tv stand, as well as coherent object attributes, \eg, the corner table. }
   \label{fig:res_zero_comp}
\end{figure*}

\begin{figure*}[ht]
  \centering
   \includegraphics[width=\linewidth]{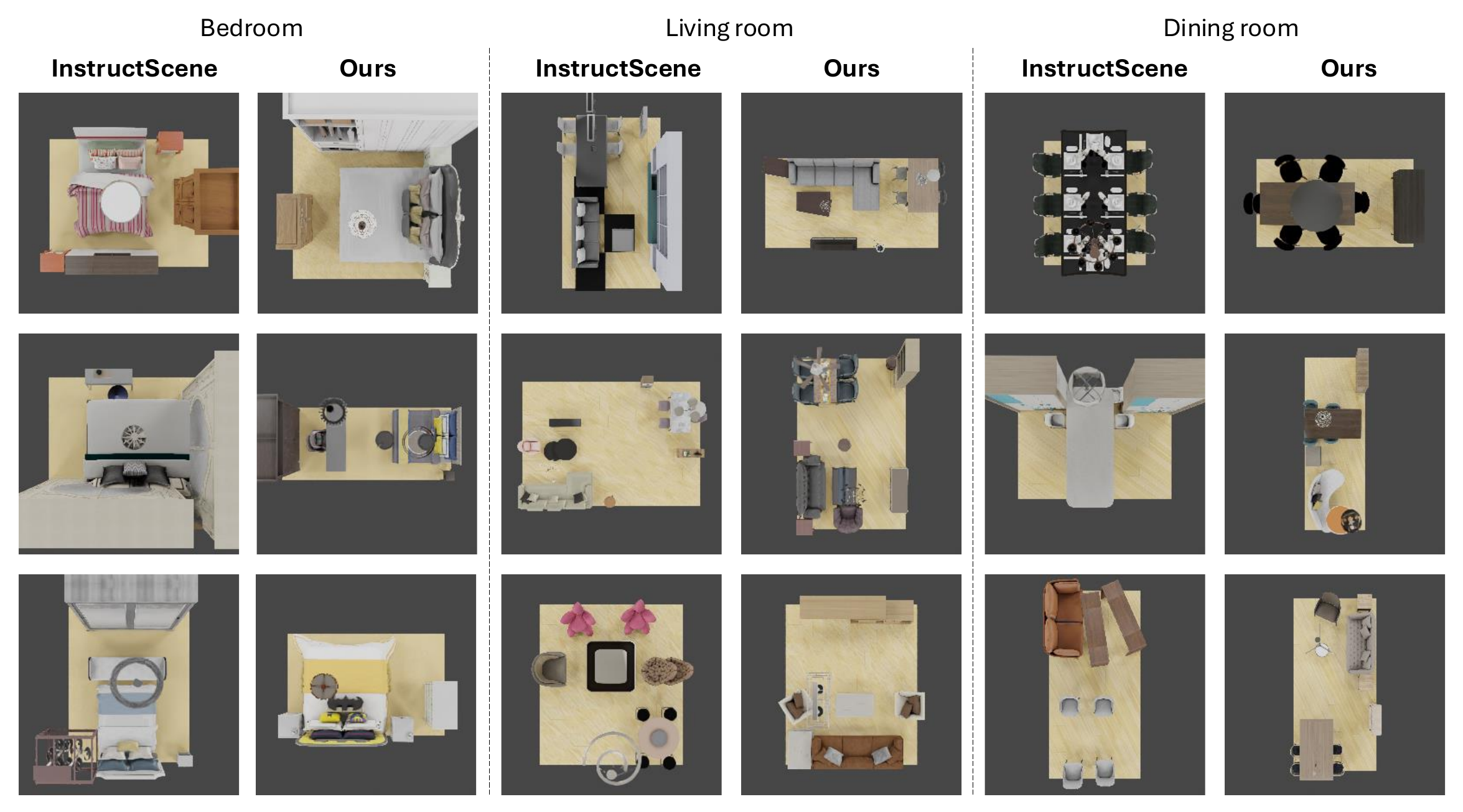}
   \caption{Unconditional generation results. Without any text-based conditions, 3D scenes of the desired room type are generated using the learned feature distributions. Ours preserves functionality and ensures practical applicability.}
   \label{fig:res_zero_unco}
\end{figure*}

\noindent\textbf{Zero-shot applications.} Qualitative examples are reported in Fig. \ref{fig:res_zero_styl} (stylization), Fig. \ref{fig:res_zero_rear} (re-arrangement), Fig. \ref{fig:res_zero_comp} (completion) and Fig. \ref{fig:res_zero_unco} (unconditional generation). We point out how InstructScene can provide content diverging from the given prompt, resulting in overlapping objects and incorrect functional object groups. For example, in the bottom row of Fig. \ref{fig:res_zero_comp}, there is no space to walk between the coffee table and the sofa. Additionally, in the last dining room of Fig. \ref{fig:res_zero_unco}, the dining table is not near dining chairs but the sofa. In contrast, the results of \modelname{} exhibit higher consistency with the input text prompts (e.g., Fig. \ref{fig:res_zero_styl}), more functional arrangements (e.g., Fig. \ref{fig:res_zero_rear}), and less object overlapping (e.g., Fig. \ref{fig:res_zero_comp}, Fig. \ref{fig:res_zero_unco}).

\end{document}